\documentclass[10pt,onecolumn]{article}

\usepackage{cvpr}

\usepackage[cmex10]{amsmath}
\usepackage{times,graphicx,amssymb,balance,url}
\usepackage{booktabs,multirow,xspace}
\usepackage[bf,footnotesize]{caption}
\usepackage{subcaption}
\usepackage{tabularx}
\usepackage[]{verbatim,bibspacing}
\usepackage[pagebackref=false,breaklinks=true,letterpaper=true,colorlinks,bookmarks=false]{hyperref}

\graphicspath{{./}{./figures/}}

\def\Vec#1{{\boldsymbol{#1}}}
\def\Mat#1{{\boldsymbol{#1}}}

\def\SPD#1{\mathcal{S}_{++}^{#1}}

\newcommand{\tr}{\mathop{\mathrm{Tr}}\nolimits}
\newcommand{\ldet}{\mathop{\mathrm{logdet}}\nolimits}

\usepackage{amsthm} 
 
\newtheorem{theorem}{Theorem}[section]

\newtheorem{definition}[theorem]{Definition}

\usepackage{authblk}

\cvprfinalcopy 

\def\cvprPaperID{} 

\ifcvprfinal\fi
\begin{document}

\title{Bregman Divergences for Infinite Dimensional Covariance Matrices}

\author[1, 2]{Mehrtash Harandi}
\author[2]{Mathieu Salzmann}
\author[1, 2]{Fatih Porikli}
\affil[1]{College of Engineering and Computer Science, Australian National University, Australia}
\affil[2]{NICTA, Canberra Research Laboratory, Australia
\thanks{NICTA is funded by the Australian Government as represented by the Department of Broadband, 
Communications and the Digital Economy and the ARC through the ICT Centre of Excellence program.
Mehrtash Harandi is partially supported by an ARC discovery grant DP130104567.}}
\affil[ ]{\tt\small {\{mehrtash.harandi,mathieu.salzmann,fatih.porikli\}@nicta.com.au}}

\maketitle

\begin{abstract}

We introduce an approach to computing and comparing Covariance Descriptors (CovDs) in infinite-dimensional spaces. CovDs have become increasingly popular to address classification problems in computer vision. 
While CovDs offer some robustness to measurement variations, they also
throw away part of the information contained in the original data by only retaining the second-order
statistics over the measurements. Here, we propose to overcome this limitation by first mapping the original data to a high-dimensional Hilbert space, and only then compute the CovDs. We show that several Bregman divergences can be computed between the resulting CovDs in Hilbert space via the use of kernels. We then exploit these divergences for classification purpose. Our experiments demonstrate the benefits of our approach on several tasks, such as material and texture recognition, person re-identification, and action recognition from motion capture data.
\end{abstract}

\section{Introduction}
\label{sec:introduction}

In this paper, we tackle the problem of employing infinite-dimensional Covariance Descriptors (CovDs) for classification.
CovDs are becoming increasingly popular in many computer vision tasks due to their robustness to measurement variations~\cite{Tuzel_PAMI_2008}. Such descriptors take the form of, \eg, region covariance matrices for pedestrian detection~\cite{Tuzel_PAMI_2008} and texture categorization~\cite{Harandi_ECCV_2012}, human joint covariances for activity recognition~\cite{Husse_IJCAI_2013}, and covariance matrices of the local Brownian motion of water molecules in diffusion tensor imaging (DTI)~\cite{Pennec_IJCV_2006}. 

As the name implies, CovDs are obtained by computing the second order statistics of feature vectors extracted at a finite number of observation points, such as the pixels of an image.
The resulting descriptors are Symmetric Positive Definite (SPD) matrices and naturally lie on non-linear manifolds known as tensor, or SPD manifolds.
As a consequence, Euclidean geometry is often not appropriate to analyze CovDs~\cite{Pennec_IJCV_2006}.
To overcome the drawbacks of Euclidean geometry and better account for the Riemannian structure of CovDs, state-of-the-art methods make use of non-Euclidean metrics (\eg,~\cite{Pennec_IJCV_2006,Sadeep_CVPR_2013}). In particular, Bregman divergences have recently been successfully employed in a number of CovD-based applications~\cite{Vemuri_CVPR2004,Sra_NIPS_2012,Harandi_ECCV_2012,Cherian_PAMI12,TSC_PAMI_2013}.

Nevertheless, all previous studies work with relatively small CovDs (\ie, at most $50 \times 50$, to the best of our knowledge) built from feature vectors whose dimension is typically much smaller than the number of observations. While this could be thought of as a filtering operation, it also implies that the information encoded in such a CovD is inherently poorer than the information jointly contained in all the observations. Recently, it was shown that CovDs could be mapped to Reproducing Kernel Hilbert Space (RKHS) via the use of SPD-specific kernels~\cite{Harandi_ECCV_2012,Sadeep_CVPR_2013}. While this may, to some degree, enhance the discriminative power of the low-dimensional CovDs, it is unlikely to be sufficient to entirely recover the information lost when constructing them.

In this paper, we overcome this issue by introducing an approach to building and analyzing infinite-dimensional CovDs from a finite number of observations. To this end, we map the original features to RKHS and compute CovDs in the resulting space. Since the dimensionality of the RKHS is much larger than the dimensionality of the observations, the resulting descriptor will encode more information than a CovD constructed in the original lower-dimensional space, and is therefore better suited for classification.

In practice, of course, the mapping to RKHS is unknown and the CovDs cannot be explicitly computed. However, here, we show that several Bregman divergences can be derived in Hilbert space via the use of kernels, thus alleviating the need for the explicit mapping. In particular, we consider the Burg~\cite{Sra_NIPS_2012}, Jeffreys~\cite{Vemuri_CVPR2004} and Stein~\cite{Sra_NIPS_2012} divergences, 
that have proven powerful to analyze 
SPD matrices. These divergences allow us to perform classification in Hilbert space via a simple nearest-neighbor (NN) classifier, or by making use of more sophisticated distance-based classifiers, such as support vector machines (SVM) with a Gaussian kernel.

We evaluated the resulting descriptors on the tasks of image-based material, texture and virus recognition, person re-identification, and action recognition from motion capture data. Our experimental evaluation clearly evidences the importance of keeping all the data information by mapping to Hilbert space before computing the CovDs. Furthermore, our empirical results show that, with this new representation, a simple NN classifier can achieve accuracies comparable to those of much more sophisticated methods, and that these accuracies can even be boosted beyond the state-of-the-art when using more powerful classifiers.

\section{Theory of Bregman Divergences}
\label{sec:preliminaries}

In this section, we review several Bregman divergences and discuss the properties that motivated our decision to use them to compare CovDs in RKHS.

Throughout the paper, we use bold upper-case letters to denote matrices (\eg, $\Mat{C}$) and bold lower-case letters for column vectors 
(\eg, $\Vec{x}$).
The $n \times n$ identity matrix is written as $\mathbf{I}_n$. 
$GL(n)$ denotes the general linear group, \ie, the group of real invertible $n \times n$ matrices.
$\SPD{n}$ is the space of $n \times n$ symmetric positive definite matrices, \ie, 
$\Mat{C} \in \SPD{n}~\mathrm{iff}~\Vec{a}^T\Mat{C}\Vec{a} >0, \forall \Vec{a} \in \mathbb{R}^n \setminus \{\Vec{0}\}$.

\begin{definition} \label{def:bregman_divergence}
		Let $\zeta : \mathcal{S}_{++}^{n} \rightarrow \mathbb{R}$ be a strictly
		convex and differentiable function defined on the symmetric positive cone $\SPD{n}$. The Bregman 
		matrix divergence 	$d_\zeta : \SPD{n} \times \SPD{n} \rightarrow [0,\infty)$ is defined as
		\begin{equation}
			\label{eqn:Bregman_Div}
			d_\zeta(\Mat{C}_1,\Mat{C}_2) = \zeta(\Mat{C}_1) - \zeta(\Mat{C}_2) - \langle \nabla_{\Mat{C}_2}\zeta , 
			\Mat{C}_1 - \Mat{C}_2 \rangle \; ,
		\end{equation}
		where
		\mbox{\small{$\langle \Mat{A} , \Mat{B} \rangle \mbox{=} \tr \left( \Mat{A}^T \Mat{B} \right) $}},
		and	$ \nabla_{\Mat{C}_2}\zeta$ is the gradient of $\zeta$ evaluated at $\Mat{C}_2$.
		The Bregman divergence is non-negative and definite (\ie, $d_\zeta(\Mat{C}_1,\Mat{C}_2) = 0 \; \text{iff}~ \Mat{C}_1 = \Mat{C}_2$).
\end{definition}

\begin{definition} \label{def:frob_norm}
	The Euclidean (Frobenius) distance is obtained by using $\zeta(\Mat{C}) =  \tr (\Mat{C}^T \Mat{C})$ as seed function in the Bregman divergence of Eq.~\ref{eqn:Bregman_Div}.
\end{definition}

\begin{definition} \label{def:Burg_divergence}
	The Burg, or $B$-, divergence is obtained by using $\zeta(\Mat{C}) = -\ldet(\Mat{C})$ 
	as seed function in the Bregman divergence of Eq.~\ref{eqn:Bregman_Div}, where $\det(\cdot)$ denotes the determinant of a matrix. 
	The B-divergence can be expressed as
	\begin{align}
		B(\Mat{C}_1,\Mat{C}_2) &=
		 \tr (\Mat{C}_1\Mat{C}_2^{-1}) -\ldet\big(\Mat{C}_1\Mat{C}_2^{-1}\big) -n\;.
		\label{eqn:Burg_Div}
	\end{align}
\end{definition}

While Bregman divergences exhibit a number of useful properties~\cite{Kulis:2009:JMLR}, their general asymmetric behavior is often counter-intuitive and undesirable in practical applications. Therefore, here, we also consider two symmetrized Bregman divergences, namely the {\it Jeffreys} and the {\it Stein} divergences.

\begin{definition} \label{def:kl_divergence}
	The Jeffreys, or $J$-, divergence 
	is obtained from the Burg divergence, and can be expressed as
	\begin{align}
		J(\Mat{C}_1,\Mat{C}_2) &= \frac{1}{2} B(\Mat{C}_1,\Mat{C}_2) + \frac{1}{2} B(\Mat{C}_2,\Mat{C}_1) \notag\\
		&= \frac{1}{2}\tr (\Mat{C}_1\Mat{C}_2^{-1}) - \frac{1}{2}\ldet \big(\Mat{C}_1\Mat{C}_2^{-1}\big) \notag\\
		&+   \frac{1}{2}\tr (\Mat{C}_2\Mat{C}_1^{-1}) - \frac{1}{2}\ldet \big(\Mat{C}_2\Mat{C}_1^{-1}\big) -n \notag\\			  
		&= \frac{1}{2}\tr (\Mat{C}_1\Mat{C}_2^{-1}) + \frac{1}{2}\tr (\Mat{C}_2\Mat{C}_1^{-1}) -n \;.
		\label{eqn:KL_Div}
	\end{align}

\end{definition}

\begin{definition} \label{def:stein_divergence}
	The Stein, or $S$-, divergence (also known as the Jensen-Bregman LogDet divergence~\cite{Cherian_PAMI12})
	 is also obtained from the Burg divergence, but through {\it Jensen-Shannon} symmetrization. It can be written as
	\begin{align}
	   	S(\Mat{C}_1,\Mat{C}_2) &= \frac{1}{2}	
   		B\hspace{-0.5ex}\left( \Mat{C}_1,
   		\frac{\Mat{C}_1\hspace{-0.5ex}+\hspace{-0.5ex}\Mat{C}_2}{2}\hspace{-0.5ex} \right)\hspace{-0.5ex} +\hspace{-0.5ex}
	   	\frac{1}{2} B\hspace{-0.5ex}\left( \Mat{C}_2,
	   	\frac{\Mat{C}_1\hspace{-0.5ex}+\hspace{-0.5ex}\Mat{C}_2}{2}\hspace{-0.5ex} \right) \notag\\
    	&= \ldet\hspace{-0.5ex} \bigg(\hspace{-0.5ex} \frac{\Mat{C}_1\hspace{-0.5ex}+\hspace{-0.5ex}\Mat{C}_2}{2}\hspace{-0.5ex} \bigg)
    	\hspace{-0.5ex}-\hspace{-0.5ex} \frac{1}{2}  \ldet\hspace{-0.5ex} \big( \Mat{C}_1\Mat{C}_2 \big). 
    	\label{eqn:Stein_Div}
	\end{align}%
\end{definition}
	
{
\renewcommand{\arraystretch}{1.2}
\begin{table*}[t]
\vspace{-0.2cm}
\centering
\begin{tabular}{| >{\centering\arraybackslash}m{2.0cm} | >{\centering\arraybackslash}m{10cm} | >{\centering\arraybackslash}m{1.5cm} | >{\centering\arraybackslash}m{2.0cm} |}
\hline    
Divergence Name & Formula & Invariance & P.D. Gaussian Kernel
\tabularnewline
\hline   
\textbf{Frobenius} & $\big\|\Mat{C}_1 - \Mat{C}_2 \big\|_F^2$ & Rotation & Yes 
\tabularnewline
\textbf{Burg} & $ \tr (\Mat{C}_1\Mat{C}_2^{-1}) -\ldet \big(\Mat{C}_1\Mat{C}_2^{-1}\big)  -n$ & Affine & No
\tabularnewline
\textbf{Jeffreys} & $\frac{1}{2}\tr\big(\Mat{C}_1\Mat{C}_2^{-1}+\Mat{C}_2\Mat{C}_1^{-1} \big) -n$ & Affine & Yes
\tabularnewline
\textbf{Stein} & $\ldet\big(\frac{1}{2}\Mat{C}_1+\frac{1}{2}\Mat{C}_2\big) -\frac{1}{2}\ldet\big(\Mat{C}_1\Mat{C}_2\big)$ & Affine & Partial
\tabularnewline
\hline
\end{tabular}
\caption{Properties of several Bregman divergences on $\SPD{n}$.} 
\label{tbl:divergences}
\end{table*}
}

\subsection{Properties of Bregman divergences}
\label{sec:subsec_bregman_prop}

Here, we present the properties of Bregman divergences that make them a natural choice as a measure of dissimilarity between two CovDs. In particular, we discuss these properties in comparison to the popular Affine Invariant Riemannian Metric (AIRM) on 
$\SPD{n}$~\cite{Pennec_IJCV_2006}, which was introduced as a geometrically-motivated way to analyze CovDs.

\subsubsection*{\textbf{Invariance to affine transformations:}}
As indicated by the name, the AIRM was designed to be invariant to affine transformations, which often is an attractive property in computer vision algorithms.
In our case, the $B$-divergence exhibits the same invariance property. More specifically, given $\Mat{A} \in \rm{GL}(n)$, 
$B(\Mat{C}_1,\Mat{C}_2) = B(\Mat{A}\Mat{C}_1\Mat{A}^T,\Mat{A}\Mat{C}_2\Mat{A}^T)$.
This can easily be shown from the definition of the $B$-divergence. 
Since the $J$- and $S$-divergences are obtained from the $B$-divergence, it can easily be verified that they inherit this affine invariance property. 
Furthermore, these two divergences are also invariant to inversion, \ie, 
\begin{align*}
	J(\Mat{C}_1,\Mat{C}_2) & = J(\Mat{C}_1^{-1},\Mat{C}_2^{-1}) \\
	S(\Mat{C}_1,\Mat{C}_2) & = S(\Mat{C}_1^{-1},\Mat{C}_2^{-1}).
\end{align*}
Finally, we also note that $B(\Mat{C}_1,\Mat{C}_2) = B(\Mat{C}_2^{-1},\Mat{C}_1^{-1})$.

\subsubsection*{\textbf{Positive definite Gaussian kernel:}}

Recently, kernel methods have been successfully employed on Riemannian manifolds~\cite{Harandi_ECCV_2012,Sadeep_CVPR_2013}. In particular, an attractive solution is to form a kernel by replacing the Euclidean distance in the popular Gaussian kernel with a more accurate metric on the manifold. However, the resulting kernel is not necessarily positive definite for any metric. In particular, the AIRM does not yield a positive definite Gaussian kernel in general. In contrast, both the $J$- and the $S$-divergences admit a Hilbert space embedding via a Gaussian kernel.
 
More specifically, for the $J$-divergence, it was shown in~\cite{Hein_2005} that the kernel 
\begin{equation}
	k_J(\Mat{C}_1,\Mat{C}_2) = \exp \{-\beta J(\Mat{C}_1,\Mat{C}_2) \},
	\label{eqn:kernel_j_div}
\end{equation}
is Conditionally Positive Definite (CPD). CPD kernels correspond to Hilbertian metrics and can be exploited in a wide range of machine learning algorithms. An example of this is kernel SVM, whose optimal solution was shown to only depend on the Hilbertian property of the metric~\cite{Hein_2005}. Note that while the kernel $k_J(\cdot,\cdot)$ was claimed to be positive definite~\cite{Moreno_2003}, we are not aware of any formal proof of this claim. 

For the $S$-divergence, the kernel 
\begin{equation}
	k_S(\Mat{C}_1,\Mat{C}_2) = \exp \{ -\beta S(\Mat{C}_1,\Mat{C}_2) \},
	\label{eqn:kernel_s_div}
\end{equation}
is not positive definite for all $\beta > 0$. However, as was shown in~\cite{Sra_NIPS_2012}, $k_S(\cdot,\cdot)$ is positive definite iff
\begin{equation*}
    \beta \in \left \{ \frac{1}{2},\frac{2}{2}, \cdots, \frac{n-1}{2} \right \}
    \cup \left \{\tau \in \mathbb{R}: \tau > \frac{1}{2}(n-1) \right \}.
\end{equation*}

Note that, here, we are not directly interested in positive definite Gaussian kernels on $\SPD{n}$ to derive our infinite-dimensional CovDs, but only to learn a kernel-based classifier with the divergences between our infinite-dimensional CovDs as input.
The properties of the Bregman divergences that we use in the remainder of this paper are summarized in Table~\ref{tbl:divergences}.

\section{Covariance Descriptors in RKHS}
\label{sec:covd_rkhs}

In this section, we show how CovDs can be computed in infinite-dimensional spaces. To this end, we first review some basics on Hilbert spaces.
\begin{definition} \label{def:hilbert_space}
A Hilbert space is a (possibly infinite-dimensional) inner product space which is complete 
with respect to the norm induced by the inner product. 
\end{definition}
An RKHS is a special type of Hilbert space with the additional property that
the inner product can be defined by a bivariate function known as the \emph{reproducing kernel}. 
For an RKHS $\big( \mathcal{H},\langle \cdot,\cdot \rangle_\mathcal{H} \big)$ on a non-empty set $\mathcal{X}$ with
$\phi:\mathcal{X} \to \mathcal{H}$
there exists a kernel function 
$k: \mathcal{X} \times \mathcal{X} \to \mathbb{R}$ such that $k(\Vec{x},\Vec{y}) = \langle \phi(\Vec{x}),\phi(\Vec{y})\rangle_\mathcal{H},~\forall \Vec{x},\Vec{y} \in \mathcal{X}$.
The concept of reproducing kernel is typically employed to recast algorithms that only exploit
inner products to high-dimensional spaces (\eg, SVM). 

Given these definitions, we now turn to the problem of computing a covariance matrix in an RKHS. Let \mbox{$\Mat{X} = \big[\Vec{x}_1 | \Vec{x}_2 |\cdots|\Vec{x}_m\big]$} be  
an ${n \times m}$ matrix, obtained by stacking $m$ independent observations
$\Vec{x}_i \in \mathbb{R}^n$ from an image or a video.
The covariance descriptor $\Mat{C} \in \SPD{n}$ is defined as
\begin{equation}	
	\Mat{C} = \dfrac{1}{m} \sum_{i=1}^{m} \big(\Vec{x}_i -\Vec{\mu} \big) \big(\Vec{x}_i -\Vec{\mu} \big)^T = \Mat{X}  \Mat{J} \Mat{J}^T \Mat{X}^T\;,	
	\label{eqn:COVD}
\end{equation}
where {\small $\Vec{\mu} = \dfrac{1}{m} \sum_{i=1}^{m} \Vec{x}_i$} is the mean of the observations, 
$\Mat{J} = {m}^{-3/2}(m\mathbf{I}_{m} - \Mat{1}_{m \times m})$ 
is a centering matrix, and
$\Mat{1}_{m \times m}$ is a square matrix with all elements equal to 1.

Let $\phi:\mathbb{R}^n \rightarrow \mathcal{H}$ be a mapping to an RKHS
whose corresponding Hilbert space $\mathcal{H}$ has dimensionality $|\mathcal{H}|$ ($|\mathcal{H}|$ could go to $\infty$).
Following Eq.~\ref{eqn:COVD}, a CovD in this RKHS can be written as
\begin{equation}
	\Mat{C}_\Mat{X} = \Phi_\Mat{X} \Mat{J} \Mat{J}^T \Phi_\Mat{X}^T\;,
	\label{eqn:COV_RKHS}
\end{equation}
where $ \Phi_\Mat{X} = \big[\phi(\Vec{x}_1) | \phi(\Vec{x}_2) |\cdots|\phi(\Vec{x}_m)\big]$.
If $|\mathcal{H}| > m$, then $\Mat{C}_\Mat{X}$ is  rank-deficient, which would make any divergence derived from the Burg divergence indefinite. More precisely, the resulting matrix would be on the boundary of the positive cone, which would make it at an infinite distance from any positive definite matrix, not only for Burg-based divergences, but also according to the AIRM.

Here, we address this issue by exploiting ideas developed in the context of covariance matrix estimation from a limited number of 
observations~\cite{bickel2008regularized,Zhou_PAMI_2006}. 
More specifically, we seek to keep the positive eigenvalues of $\Mat{C}_\Mat{X}$ intact and replace the zero ones with a very small positive number $\rho$, thus making the CovD positive definite. 
First, using a standard result~\cite{Turk_1991}, we note that the positive eigenvalues of $\Mat{C}_\Mat{X}$, 
denoted by $\Lambda_{\Mat{X}}$, can be computed 
from $\Mat{J}^T\Phi_{\Mat{X}}^T\Phi_{\Mat{X}}\Mat{J} = \Mat{J}^T\Mat{K}_{\Mat{X},\Mat{X}}\Mat{J}$, where $\Mat{K}_{\Mat{X},\Mat{X}}$ is the $m \times m$ kernel matrix whose elements are defined by the kernel function
$k(\Vec{x}_i,\Vec{x}_j)$. By eigenvalue decomposition, we can write
\begin{equation}
	\Mat{J}^T\Mat{K}_{\Mat{X},\Mat{X}}\Mat{J} = \Mat{V}_{\Mat{X}} \Lambda_{\Mat{X}}\Mat{V}_{\Mat{X}}^T\;.
	\label{eqn:svd_kXX}
\end{equation}
This lets us write a (regularized) estimate of $\Mat{C}_\Mat{X}$ as
\begin{equation}
\widehat{\Mat{C}}_\Mat{X} = \Phi_{\Mat{X}} \Mat{W}_{\Mat{X}} \Mat{W}_{\Mat{X}}^T \Phi_{\Mat{X}}^T + \rho \mathbf{I}_{|\mathcal{H}|}\;,
\label{eqn:est_cov_matrix}
\end{equation}
where
\begin{equation}
	\Mat{W}_{\Mat{X}} = \Mat{J} \Mat{V}_{\Mat{X}}\left( \mathbf{I}_{\Mat{X}} - \rho \Lambda_{\Mat{X}}^{-1} \right)^{\frac{1}{2}}\;,
	\label{eqn:w_x}
\end{equation}
with $\mathbf{I}_{\Mat{X}}$ the identity matrix whose dimension is the number of positive eigenvalues of $\Mat{C}_\Mat{X}$
~\cite{Zhou_PAMI_2006}. 
Note that this derivation can also be employed to model points in $\mathcal{H}$ with lower-dimensional latent variables by retaining only the top $r$ eigenvalues and eigenvectors of 
$\Mat{J}^T\Mat{K}_{\Mat{X},\Mat{X}}\Mat{J}$ to form $\Mat{W}_{\Mat{X}}$~\cite{Bishop_2006}.

\section{Bregman Divergences in RKHS}
\label{sec:bregman_divergence_rkhs}

In this section, we derive different Bregman divergences for the infinite-dimensional CovDs introduced in Section~\ref{sec:covd_rkhs}.  
In these derivations, we will make use of the equivalence
\begin{equation}
	\Mat{W}_{\Mat{X}}^T \Phi_{\Mat{X}}^T \Phi_{\Mat{X}} \Mat{W}_{\Mat{X}}  
	= \Lambda_{\Mat{X}} - \rho \mathbf{I}_\Mat{X}\;,
	\label{eqn:wkwt}
\end{equation}
whose derivation is provided in supplementary material.

\subsubsection*{\textbf{Euclidean Metric:}}

The Frobenius norm can easily be computed as 
\begin{align*}
	\delta_e^2(\widehat{\Mat{C}}_\Mat{X},\widehat{\Mat{C}}_\Mat{Y}) &= 
	\big \| \Phi_{\Mat{X}} \Mat{W}_{\Mat{X}} \Mat{W}_{\Mat{X}}^T \Phi_{\Mat{X}}^T - 
	\Phi_{\Mat{Y}} \Mat{W}_{\Mat{Y}} \Mat{W}_{\Mat{Y}}^T \Phi_{\Mat{Y}}^T \big \|_F^2  \notag \\
	&\hspace{-2.3cm}=  \big{\|}\Lambda_\Mat{X} - \rho \mathbf{I}_\Mat{X}\big{\|}_F^2 +   
	\big{\|}\Lambda_\Mat{Y} - \rho \mathbf{I}_\Mat{Y}\big{\|}_F^2 -2\big{\|}\Mat{W}_{\Mat{Y}}^T \Mat{K}_{\Mat{Y},\Mat{X}}\Mat{W}_{\Mat{X}}\big{\|}_F^2 .		
\end{align*}
Note that, although not a desirable property~\cite{Pennec_IJCV_2006}, the Euclidean metric is definite for positive semi-definite matrices, which makes it possible
to set $\rho$ to zero. 

\subsubsection*{\textbf{Burg Divergence:}}

Using the Sylvester determinant theorem~\cite{Golub_Book}, we first note that
\begin{align}
	\det \big( \widehat{\Mat{C}}_{\Mat{X}}\big) &= \det \big(\Phi_{\Mat{X}} \Mat{W}_{\Mat{X}} \Mat{W}_{\Mat{X}}^T \Phi_{\Mat{X}}^T
	 + \rho \mathbf{I}_\mathcal{|H|} \big) \notag \\
	&= \rho^\mathcal{|H|} \det \big(\mathbf{I}_{\Mat{X}} + 
	\frac{1}{\rho} \Mat{W}_{\Mat{X}}^T \Phi_{\Mat{X}}^T \Phi_{\Mat{X}} \Mat{W}_{\Mat{X}}\big) \notag \\
	&= \rho^\mathcal{|H|} \det \big(\mathbf{I}_{\Mat{X}} + \frac{1}{\rho} \left( \Lambda_{\Mat{X}} - 
	\rho \mathbf{I}_{\Mat{X}} \right) \big) \notag \\
	&= \rho^{\mathcal{|H|}} \det \big( \rho^{-1} \Lambda_{\Mat{X}} \big).	
	\label{eqn:det_cov_rkhs}
\end{align}

Then, from the Woodbury matrix identity~\cite{Golub_Book}, we have
\begin{align}
	\widehat{\Mat{C}}^{-1}_{\Mat{Y}} &= \Big( \Phi_{\Mat{Y}} \Mat{W}_{\Mat{Y}} \Mat{W}_{\Mat{Y}}^T \Phi_{\Mat{Y}}^T + 
	\rho \mathbf{I}_\mathcal{|H|} \Big)^{-1} \notag \\
	&= \frac{1}{\rho} \mathbf{I}_\mathcal{|H|} - \frac{1}{\rho} \Phi_{\Mat{Y}} \Mat{W}_{\Mat{Y}} 
	\Lambda_\Mat{Y}^{-1} \Mat{W}_{\Mat{Y}}^T \Phi_{\Mat{Y}}^T.
	\label{eqn:p_inv_CX}
\end{align}

This lets us write,
\begin{align}
	&\tr\Big(\widehat{\Mat{C}}_{\Mat{X}}\widehat{\Mat{C}}^{-1}_{\Mat{Y}}\Big) =		
	\mathcal{|H|} +\hspace{-0.5ex} \tr \big(\dfrac{1}{\rho} \Lambda_{\Mat{X}} - \mathbf{I}_\Mat{X} \big)
	 \hspace{-0.5ex}-\hspace{-0.5ex}\tr \big( \mathbf{I}_\Mat{Y}  - \rho \Lambda_{\Mat{Y}}^{-1} \big)
	 \notag\\
	 &-\frac{1}{\rho}\tr \Big( \Mat{W}_{\Mat{X}}^T \Mat{K}_{\Mat{X},\Mat{Y}} \Mat{W}_{\Mat{Y}} \Lambda_\Mat{Y}^{-1} \Mat{W}_{\Mat{Y}}^T 
	    \Mat{K}_{\Mat{Y},\Mat{X}} \Mat{W}_{\Mat{X}} \Big)	\;.
	\label{eqn:tr_Xinv_Y}
\end{align}
By combining Eqs.~\ref{eqn:det_cov_rkhs} and~\ref{eqn:tr_Xinv_Y}, we then obtain
\begin{align}
	&B_\mathcal{H}\Big(\widehat{\Mat{C}}_{\Mat{X}},\widehat{\Mat{C}}_{\Mat{Y}}\Big) =
		\tr \big( \dfrac{1}{\rho}\Lambda_{\Mat{X}} - \mathbf{I}_\Mat{X} \big)
	    -\tr \big( \mathbf{I}_\Mat{Y}  - \rho \Lambda_{\Mat{Y}}^{-1} \big) \notag \\
	    &-\frac{1}{\rho}\tr \Big( \Mat{W}_{\Mat{X}}^T \Mat{K}_{\Mat{X},\Mat{Y}} \Mat{W}_{\Mat{Y}} \Lambda_\Mat{Y}^{-1} \Mat{W}_{\Mat{Y}}^T 
	    \Mat{K}_{\Mat{Y},\Mat{X}} \Mat{W}_{\Mat{X}} \Big)	\notag \\    	     
	    &+ \ldet \Big(\rho^{-1} \Lambda_{\Mat{Y}}\Big) - \ldet \Big(\rho^{-1} \Lambda_{\Mat{X}}\Big). 
	\label{eqn:burg_rkhs}
\end{align}
Note that the Burg divergence is independent of $|\mathcal{H}|$. This property is inherited by the Jeffreys and Stein divergences derived below.

\subsubsection*{\textbf{Jeffreys Divergence:}}

From the definition in Section~\ref{sec:preliminaries}, the Jeffreys divergence can be obtained directly from the Burg divergence. This yields
\begin{align}
	&J_{\mathcal{H}}\Big(\widehat{\Mat{C}}_{\Mat{X}},\widehat{\Mat{C}}_{\Mat{Y}}\Big) =
	    \frac{1}{2\rho}\tr \big( \Lambda_{\Mat{X}} - \rho \mathbf{I}_\Mat{X} \big) + 
	    \frac{1}{2\rho}\tr \big( \Lambda_{\Mat{Y}} - \rho \mathbf{I}_\Mat{Y} \big)  \notag\\
	    &-\frac{1}{2\rho}\tr \Big( \Mat{W}_{\Mat{X}}^T \Mat{K}_{\Mat{X},\Mat{Y}} \Mat{W}_{\Mat{Y}} \Lambda_\Mat{Y}^{-1} \Mat{W}_{\Mat{Y}}^T 
	    \Mat{K}_{\Mat{Y},\Mat{X}} \Mat{W}_{\Mat{X}} \Big)  \notag\\
	    &-\frac{1}{2\rho}\tr \Big( \Mat{W}_{\Mat{Y}}^T \Mat{K}_{\Mat{Y},\Mat{X}} \Mat{W}_{\Mat{X}} \Lambda_\Mat{X}^{-1} \Mat{W}_{\Mat{X}}^T 
	    \Mat{K}_{\Mat{X},\Mat{Y}} \Mat{W}_{\Mat{Y}} \Big)  \notag\\
	    &-\frac{1}{2}\tr \big( \mathbf{I}_\Mat{X}  - \rho \Lambda_{\Mat{X}}^{-1} \big)  -
	    \frac{1}{2}\tr \big( \mathbf{I}_\Mat{Y}  - \rho \Lambda_{\Mat{Y}}^{-1} \big)\; . 
	\label{eqn:jefferys_rkhs}
\end{align}

\subsubsection*{\textbf{Stein Divergence:}}
\label{sec:stein_metric}

To compute the Stein divergence in $\mathcal{H}$, 
let us first define 
\begin{eqnarray}
\Mat{Q} = \left[ \begin{array}{cc} \Mat{W}_\Mat{X}  &\Mat{0} \\
\Mat{0} &\Mat{W}_\Mat{Y} \end{array} \right].
\label{eqn:Q}
\end{eqnarray}

This lets us write
\begin{equation}
	\dfrac{\widehat{\Mat{C}}_{\Mat{X}}+\widehat{\Mat{C}}_{\Mat{Y}}}{2} = \rho \mathbf{I}_{|\mathcal{H}|} + \dfrac{1}{2}
	\big[  \Phi_{\Mat{X}} \: \Phi_{\Mat{Y}} \big] \Mat{Q}	\Mat{Q}^T
	\left[ \begin{array}{c} \Phi_{\Mat{X}}^T\vspace{1ex} \\  \Phi_{\Mat{Y}}^T \end{array} \right].	
	\label{eqn:CX_CY}
\end{equation}

Similarly as in Eq.~\ref{eqn:det_cov_rkhs}, $\det\big( (\widehat{\Mat{C}}_{\Mat{X}} + \widehat{\Mat{C}}_{\Mat{Y}})/2 \big)$ becomes
\begin{align*}	
	&\rho^{\mathcal{|H|}}\det \Bigg( \mathbf{I}_{|\mathcal{H}|} + \dfrac{1}{2\rho}\big[  \Phi_{\Mat{X}} \: \Phi_{\Mat{Y}} \big] 
	\Mat{Q}\Mat{Q}^{T}	
	\left[ \begin{array}{c} \Phi_{\Mat{X}}^T \vspace{1ex}\\ \Phi_{\Mat{Y}}^T \end{array} \right]  \Bigg) \\
	= &\rho^{\mathcal{|H|}}\det \Bigg(\mathbf{I}_{\Mat{X} + \Mat{Y}} + \dfrac{1}{2\rho}\Mat{Q}^{T}	
	\left[ \begin{array}{c} \Phi_{\Mat{X}}^T \vspace{1ex} \\ \Phi_{\Mat{Y}}^T \end{array} \right]  
	\big[  \Phi_{\Mat{X}} \: \Phi_{\Mat{Y}} \big] \Mat{Q}\Bigg) \\
	= &\rho^{\mathcal{|H|}} \det \Bigg(\mathbf{I}_{\Mat{X} + \Mat{Y}} + \dfrac{1}{2\rho}\Mat{Q}^{T}
	\mathbb{K}_{\Mat{X},\Mat{Y}}\Mat{Q}\Bigg)\;,
\label{eqn:Proof_CX_CY}
\end{align*}
where
\begin{equation}
	\mathbb{K}_{\Mat{X},\Mat{Y}} = \left[ \begin{array}{cc} \Mat{K}_{\Mat{X},\Mat{X}} &\Mat{K}_{\Mat{X},\Mat{Y}} \\ 
	\Mat{K}_{\Mat{Y} , \Mat{X}} &\Mat{K}_{\Mat{Y},\Mat{Y}}\end{array} \right]  \;.
	\label{eqn:big_K_stein}
\end{equation} 

Therefore, we have
\begin{align}
&S_{\mathcal{H}}\Big(\widehat{\Mat{C}}_{\Mat{X}},\widehat{\Mat{C}}_{\Mat{Y}}\Big) = 
	\ldet \Big(\mathbf{I}_{\Mat{X} + \Mat{Y}}  + \dfrac{1}{2\rho}\Mat{Q}^{T}	
	\mathbb{K}_{\Mat{X},\Mat{Y}} \Mat{Q}\Big)\nonumber\\
	&-\dfrac{1}{2}  \ldet \big( \rho^{-1}\Lambda_{\Mat{X}}\big)
	-\dfrac{1}{2}  \ldet \big( \rho^{-1}\Lambda_{\Mat{Y}}\big)\;.
	\label{eqn:stein_rkhs}
\end{align}

\subsection{Practical Considerations}

When computing divergences in RKHS, it is desirable to minimize the effect of the parameter $\rho$, and thus have divergences that do not depend on its inverse.
To this end, let us assume that 
the same number of eigenvectors were kept to build $\Mat{C}_{\Mat{X}}$ and $\Mat{C}_{\Mat{Y}}$. 
In this case, the Stein divergence can be written as
\begin{align}
	&\widehat{S}_\mathcal{H}\Big(\widehat{\Mat{C}}_{\Mat{X}},\widehat{\Mat{C}}_{\Mat{Y}}\Big) =
	\ldet{\Big( \rho\mathbf{I}_{\Mat{X} + \Mat{Y}} + \dfrac{1}{2}\Mat{Q}^{T}\mathbb{K}_{\Mat{X},\Mat{Y}}	\Mat{Q}\Big) } \notag \\
		&-\dfrac{1}{2}  \ldet{ (\Lambda_{\Mat{X}})}
		-\dfrac{1}{2}  \ldet{ (\Lambda_{\Mat{Y}})}\;,
\label{eqn:stein_rkhs_simple}
\end{align} 
where the term $\rho\mathbf{I}_{\Mat{X} + \Mat{Y}}$ can be thought of as a regularizer for 
$\dfrac{1}{2}\Mat{Q}^{T}\mathbb{K}_{\Mat{X},\Mat{Y}}\Mat{Q}$. 
For the Jeffreys divergence, we can define
\begin{align}
	&\widehat{J}_\mathcal{H}\Big(\widehat{\Mat{C}}_{\Mat{X}},\widehat{\Mat{C}}_{\Mat{Y}}\Big) =
		\lim_{\rho \to 0} 2\rho J_\mathcal{H}\Big(\widehat{\Mat{C}}_{\Mat{X}},\widehat{\Mat{C}}_{\Mat{Y}}\Big) = \notag\\	       
	    &-\tr \Big( \Mat{W}_{\Mat{X}}^T \Mat{K}_{\Mat{X},\Mat{Y}} \Mat{W}_{\Mat{Y}}
	     \Lambda_{\Mat{Y}}^{-1} \Mat{W}_{\Mat{Y}}^T 
	    \Mat{K}_{\Mat{Y},\Mat{X}} \Mat{W}_{\Mat{X}} \Big)  \notag\\
	    &-\tr \Big( \Mat{W}_{\Mat{Y}}^T \Mat{K}_{\Mat{Y},\Mat{X}} \Mat{W}_{\Mat{X}} \Lambda_\Mat{X}^{-1} 
	    \Mat{W}_{\Mat{X}}^T    \Mat{K}_{\Mat{X},\Mat{Y}} \Mat{W}_{\Mat{Y}} \Big) \notag \\
	    &+\tr ( \Lambda_{\Mat{X}}) + \tr ( \Lambda_{\Mat{Y}})\; . 
	\label{eqn:simple_jefferys_rkhs}
\end{align}
In our experiments, we used the definitions of Eqs.~\ref{eqn:simple_jefferys_rkhs} and~\ref{eqn:stein_rkhs_simple}.

\subsection{Computational Complexity}

Here we compare the complexity of computing $J_\mathcal{H}(\cdot,\cdot)$ and $S_\mathcal{H}(\cdot,\cdot)$ against that of
$J(\cdot,\cdot)$ and $S(\cdot,\cdot)$. Let $\Mat{X} \in \mathbb{R}^{n \times m}$ 
and $\Mat{Y} \in \mathbb{R}^{n \times m}$ be two given sets of observation, with $m \gg n$. 

Computing the $n \times n$ CovDs based on Eq.~\ref{eqn:COVD} 
requires $O(n^2m)$.  The inverse of an $n \times n$ SPD matrix can be computed by Cholesky decomposition
in $\frac{1}{2}n^3$ flops. Therefore, computing the $J$-divergence requires $2n^2m + 2n^{2.3} + n^3$ flops, which is dominated by $2n^2m$.
The complexity of computing the determinant of an $n \times n$ matrix by
Cholesky decomposition is $O(\frac{1}{3}n^3)$. Therefore, computing the $S$-divergence requires 
$2n^2m + n^{2.3} + \frac{2}{3}n^3$ flops, which is again dominated by $2n^2m$.

In RKHS, computing $K_{\Mat{X},\Mat{X}}$, $K_{\Mat{Y},\Mat{X}}$ and $K_{\Mat{X},\Mat{Y}}$ requires $m^2$ flops for each matrix. Therefore, evaluating Eq.~\ref{eqn:svd_kXX} requires for $m^3$ flops. Assuming that $r$, $r < m$, eigenvectors are used to create $\Mat{W}_\Mat{X}$ in Eq.~\ref{eqn:w_x}, computing $J_\mathcal{H}$ according to Eq.~\ref{eqn:jefferys_rkhs} requires 
$2m^3 + 3m^2 + 4m^2r +2mr^2$. For $S_\mathcal{H}$, evaluating Eq.~\ref{eqn:stein_rkhs} takes $6m^3 + 3m^2 + 8m^2r + 8mr^2$ flops.

Generally speaking, the complexity of computing the Jeffreys and Stein divergences in the observation space is linear in $m$ while it is cubic
when working in RKHS. Our experimental evaluation shows, however, that working in RKHS remains practical. To illustrate this, we compare the runtimes required to compute the Stein divergence between $500,000$ pairs 
of CovDs on $\SPD{10}$ using Eq.~\ref{eqn:Stein_Div} and Eq.~\ref{eqn:stein_rkhs_simple}. Each CovD on $\SPD{10}$ was obtained from $m = 100$ observations. 
In the observation space, computing the Stein divergence on an i7 machine using Matlab took 53s. For $S_{\mathcal{H}}$, it took 452s, 566s and 868s when keeping 10, 20 and 50 eigenvectors to estimate the covariances, respectively. While slower, these runtimes remain perfectly acceptable, especially when considering the large accuracy gain that working in RKHS entails, as evidenced by our experiments.

\section{Experimental Evaluation}
\label{sec:experiments}

We now present our empirical results obtained with the infinite-dimensional CovDs and their Bregman divergences defined in Sections~\ref{sec:covd_rkhs} and~\ref{sec:bregman_divergence_rkhs}. In particular, due to their symmetry and the fact that they yield valid Gaussian kernels, we utilized the Jeffreys and Stein divergences, and relied on two different classifiers for each divergence: A simple nearest neighbor classifier, which clearly evidences the benefits of using infinite-dimensional CovDs, and an SVM classifier with a Gaussian kernel, which further boosts the performance of our infinite-dimensional CovDs.

The different algorithms evaluated in our experiments are referred to as:
\begin{itemize}
	\renewcommand{\labelitemi}{\scriptsize$$}
	\vspace{-0.15cm}
	\item \textbf{$J/S$-NN:} Jeffreys/Stein based Nearest Neighbor classifier on CovDs in the observation space.	
	\vspace{-0.15cm}
	\item \textbf{$J/S$-SVM:} Jeffreys/Stein based kernel SVM on CovDs in the observation space.
	\vspace{-0.15cm}
	\item \textbf{$J_\mathcal{H}/S_\mathcal{H}$-NN:} Jeffreys/Stein based Nearest Neighbor classifier on infinite-dimensional CovDs.
	\vspace{-0.15cm}
	\item \textbf{$J_\mathcal{H}/S_\mathcal{H}$-SVM:} Jeffreys/Stein based kernel SVM on infinite-dimensional CovDs.	
\end{itemize}
We also provide the results of the PLS-based Covariance Discriminant Learning (CDL) technique of~\cite{Wang_CVPR_2012_CDL}, which can be considered as the state-of-the-art for CovD-based classification. 
In all our experiments, we used the RBF kernel to create infinite-dimensional CovDs. The parameters of our algorithm, \ie, the RBF bandwidth and the number of eigenvectors $r$,
were determined by cross-validation.

\def \VIRUS_SCALE{1.0}
\begin{figure}[!tb]
\centering
	\includegraphics[width= \VIRUS_SCALE \columnwidth,keepaspectratio]{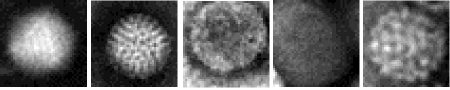}
	\caption{
    \small
        Sample images from the virus dataset~\cite{Virus_Dataset}.
    }
\label{fig:virus_dataset}
\end{figure}

\begin{table}[!tb]
 	\caption    {    
		   Recognition accuracies for the virus dataset~\cite{Virus_Dataset}.
    }
  	\centering
    \begin{tabular}{lc}
    	\toprule
    	{\bf Method} &{\bf Recognition Accuracy }\\
    	\toprule  	
    	{\bf $B$-NN}									&$56.6\% \pm 2.7$\\
    	{\bf $J$-NN}									&$60.3\% \pm 5.3$\\	
    	{\bf $S$-NN}									&$60.7\% \pm 5.4$\\
    	{\bf CDL~\cite{Wang_CVPR_2012_CDL}}    			&$69.5\% \pm 3.1$\\
    	{\bf $J$-SVM}									&$73.9\% \pm 4.0$\\
    	{\bf $S$-SVM}									&$76.5\% \pm 3.3$\\
    	\midrule
    	{\bf $B_{\mathcal{H}}$-NN}     					&$63.5\% \pm 3.4$\\
    	{\bf $J_{\mathcal{H}}$-NN}     					&$66.7\% \pm 4.2$\\
    	{\bf $S_{\mathcal{H}}$-NN}                  	&$67.1\% \pm 4.3$\\
    	\midrule
    	{\bf $J_{\mathcal{H}}$-SVM}	               		&$81.1\% \pm 3.4$\\    			
		{\bf $S_{\mathcal{H}}$-SVM}	               		&$\bf 81.2\% \pm 2.9$\\    			
		\bottomrule	
    \end{tabular}
    \label{tab:table_virus_performance}
\end{table}

\subsection{Virus Classification}
\label{sec:exp_virus}

As a first experiment, we used the virus dataset~\cite{Virus_Dataset} which contains 15 different virus classes.
Each class has 100 images of size $41 \times 41$ that were segmented automatically~\cite{Virus_Dataset}.
Samples from the virus dataset are shown in Fig.~\ref{fig:virus_dataset}. We used the 10 splits 
provided with the dataset in a leave-one-out manner, \ie, 10 experiments with 9 splits for training and 1 split as query.

At each pixel $(u,v)$ of an image, we computed the 25-dimensional feature vector
\begin{small}
\begin{equation*}
	\Vec{x}_{u,v} =	\hspace{-0.1cm}\bigg[I_{u,v}, \left| \dfrac{\partial I}  {\partial u}  \right|, 
	\left|\frac{\partial I}  {\partial v}  \right|,
    \left| \frac{\partial^2 I}{\partial u^2}\right|, \left|\frac{\partial^2 I}{\partial v^2}\right|, \big|G^{0,0}_{u,v}\big|,
    \cdots,\big|G^{4,5}_{u,v}\big| ~\bigg]^T,
\end{equation*}%
\end{small}
\noindent where $I_{u,v}$ is the intensity value, $G^{o,s}_{u,v}$ is the response of a 2D Gabor wavelet~\cite{Lee_PAMI_1996}
 with orientation $o$ and scale $s$, and $|\cdot|$ denotes the magnitude of a 
complex value. Here, we generated 20 Gabor filters at 4 orientations and 5 scales.

We report the mean recognition accuracies over the 10 runs in Table~\ref{tab:table_virus_performance}. The NN results clearly show that the CovDs computed in RKHS are more discriminative than the ones built directly from the original features. Note that applying kernel SVM boosts the performance of all the CovDs. Note also that our simple NN scheme in RKHS achieves comparable performance to the more involved CDL. Our $J_\mathcal{H}/S_\mathcal{H}$-SVM methods outperform all the baselines. Here, for each split, the runtimes were on average 130s for the Stein divergence in observation space and 1180s for $S_\mathcal{H}$, which remains perfectly practical.

In addition to the baselines in Table~\ref{tab:table_virus_performance}, we evaluated the performance of 
Local Binary Patterns (LBP)~\cite{LBP_PAMI_2002} and Gabor filters~\cite{Lee_PAMI_1996}, which are popular methods to analyze textures. 
With an NN classifier, we obtained accuracies of $36.8\% \pm 3.9$ and $33.7\% \pm 4.0$ for LBP and Gabor filters, respectively. 
This clearly shows the difficulty of this task and the notable improvement achieved by using CovDs. 

With this dataset, we also evaluated the performance of the Euclidean metric and asymmetric Burg divergence in RKHS. We obtained $52.4\%$ and $63.5\%$ accuracy for the Euclidean metric and the Burg divergence, respectively. This indicates that a simple Euclidean metric is poorly-suited to handle CovDs. In the remainder of this section, we focus on the Stein and Jeffreys divergences.

\def \KTHTIPS_SCALE {0.225}
\begin{figure}[!tb]			
	\includegraphics[width = \KTHTIPS_SCALE \columnwidth]{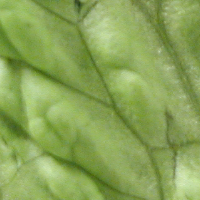}
	\includegraphics[width = \KTHTIPS_SCALE \columnwidth]{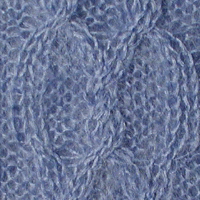}
	\includegraphics[width = \KTHTIPS_SCALE \columnwidth]{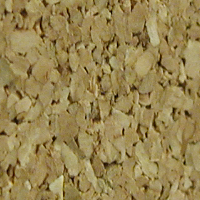}
	\includegraphics[width = \KTHTIPS_SCALE \columnwidth]{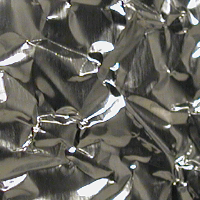}
	\caption{Samples from the KTH-TIPS2b material dataset~\cite{KTH_TIPS2_Dataset}.}
	\label{fig:KTH_TIPS_Dataset}
\end{figure}

\begin{table}[!tb]
 	\caption    {    
		Recognition accuracies for the KTH-TIPS2b material dataset~\cite{KTH_TIPS2_Dataset}.
    }
  	\centering
  	\small
    \begin{tabular}{lccccc}
    	\toprule
    	{\bf Method} &{\bf Split\#1} &{\bf Split\#2} &{\bf Split\#3} &{\bf Split\#4}	&{\bf Average}\\
    	\toprule  	
    	{\bf $J$-NN}							&$72.6\%$			&$72.8\%$			&$64.8\%$			&$64.3\%$		&$68.6\%$\\	
    	{\bf $S$-NN}							&$72.5\%$			&$73.4\%$			&$64.6\%$			&$64.6\%$		&$68.8\%$\\	
    	{\bf CDL~\cite{Wang_CVPR_2012_CDL}}    	&$83.5\%$			&$75.6\%$			&$71.5\%$			&$74.5\%$		&$76.3\%$\\	
    	{\bf $J$-SVM}							&$77.4\%$			&$76.6\%$			&$71.4\%$			&$73.3\%$		&$74.7\%$\\	
    	{\bf $S$-SVM}							&$83.6\%$			&$\bf 80.9\%$		&$73.1\%$			&$75.4\%$		&$78.3\%$\\	
    	\midrule
    	{\bf $J_{\mathcal{H}}$-NN}     			&$79.1\%$			&$75.7\%$			&$69.9\%$			&$67.7\%$		&$73.1\%$\\	
    	{\bf $S_{\mathcal{H}}$-NN}              &$78.1\%$			&$76.3\%$			&$69.2\%$			&$67.8\%$		&$72.9\%$\\	
    	\midrule
    	{\bf $J_{\mathcal{H}}$-SVM}	            &$\bf 85.2\%$		&$78.5\%$			&$\bf 76.4\%$		&$79.7\%$		&$79.9\%$\\	   			
		{\bf $S_{\mathcal{H}}$-SVM}	            &$85.1\%$			&$79.8\%$			&$74.0\%$			&$\bf 81.6\%$	&$\bf 80.1\%$\\	   			
		\bottomrule	
    \end{tabular}
    \label{tab:table_KTH_TIPS2_performance}
\end{table}		

\subsection{Material Categorization}
\label{sec:exp_material_cat}

We then used the KTH-TIPS2b dataset~\cite{KTH_TIPS2_Dataset} to perform material categorization. 
KTH-TIPS2b contains images of 11 materials captured under 4 different illuminations, in 3 poses and at 9 scales. 
This yields a total of $3 \times 4 \times 9 = 108$ images for each sample in a category, with 4 samples per material. 
We resized the original images to {$128 \times 128$} pixels and generated CovDs from 1024 observations computed on a coarse grid (\ie, every 4 pixels horizontally and vertically).
At each point on the grid, we extracted the 23-dimensional feature vector
\begin{equation*}
	\Vec{x}_{u,v} =	\bigg[r_{u,v}, ~g_{u,v}, ~b(u,v), \big|G^{0,0}_{u,v}\big|,~\cdots,~\big|G^{4,5}_{u,v}\big| ~\bigg]^T\;,
\end{equation*}%
\noindent
where $r_{u,v}$, $g_{u,v}$ and $b_{u,v}$ are the color intensities, and $G^{o,s}_{u,v}$ are the same Gabor filter responses as before.

In Table~\ref{tab:table_KTH_TIPS2_performance}, we report the recognition accuracies computed by training on 3 samples per category and testing on the remaining sample. On average, with an NN classifier, our infinite-dimensional CovDs outperform the $23\times 23$-dimensional ones by more than $4\%$. As before, kernel SVM further improves the performance of all CovDs. This yields a maximum average accuracy of 80.1\% for our {\bf $S_{\mathcal{H}}$-SVM}, which, to the best of our knowledge, is state-of-the-art on this dataset~\cite{Liu_IVC_2012}.

\def \KYLBERG_SCALE{0.225}
\begin{figure}[!tb]
\centering
	\includegraphics[width= \KYLBERG_SCALE \columnwidth,keepaspectratio]{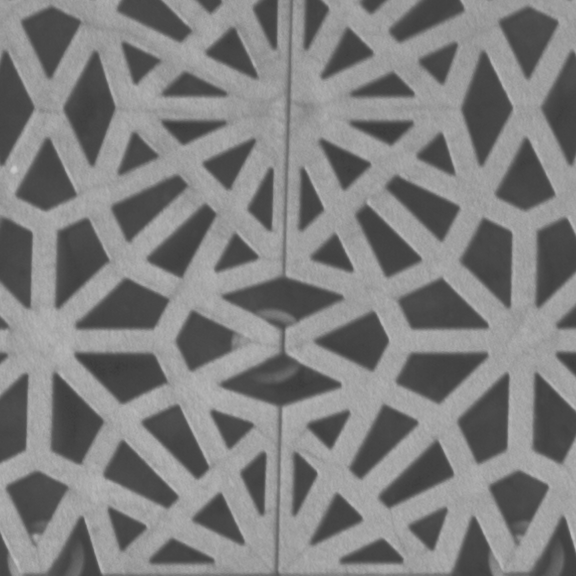}
	\includegraphics[width= \KYLBERG_SCALE \columnwidth,keepaspectratio]{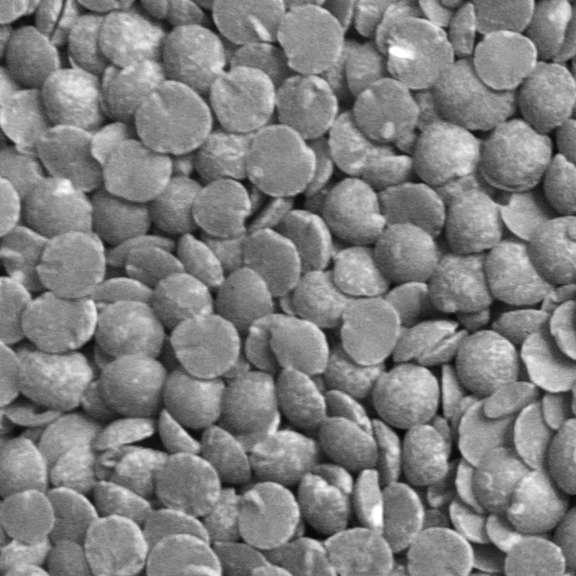}
	\includegraphics[width= \KYLBERG_SCALE \columnwidth,keepaspectratio]{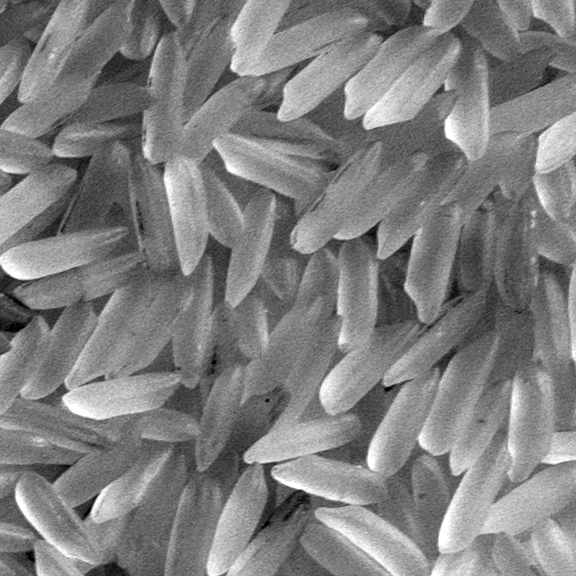}
	\includegraphics[width= \KYLBERG_SCALE \columnwidth,keepaspectratio]{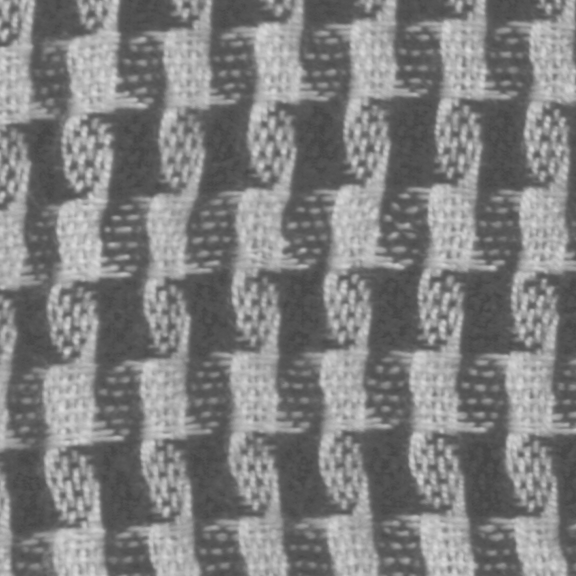}
	\caption{
    \small
        Samples from the Kylberg texture dataset~\cite{Kylberg_Dataset}.
    }
\label{fig:kylberg_dataset}
\end{figure}

\begin{table}[!tb]
 	\caption    {    
		   Recognition accuracies for the Kylberg dataset~\cite{Kylberg_Dataset}.
    }
  	\centering
    \begin{tabular}{lc}
    	\toprule
    	{\bf Method} &{\bf Recognition Accuracy }\\
    	\toprule  	
    	{\bf $J$-NN}									&$77.00\% \pm 1.40$\\	
    	{\bf $S$-NN}									&$77.02\% \pm 1.37$\\
    	{\bf CDL~\cite{Wang_CVPR_2012_CDL}}    			&$79.87\% \pm 1.06$\\
    	{\bf $J$-SVM}									&$82.19\% \pm 1.30$\\
    	{\bf $S$-SVM}									&$81.27\% \pm 1.07$\\
    	\midrule
    	{\bf $J_{\mathcal{H}}$-NN}     					&$84.89\% \pm 1.06$\\
    	{\bf $S_{\mathcal{H}}$-NN}                  	&$84.91\% \pm 1.08$\\
    	\midrule
    	{\bf $J_{\mathcal{H}}$-SVM}	               		&$91.25\% \pm 1.33$\\    			
		{\bf $S_{\mathcal{H}}$-SVM}	               		&$\bf 91.36\% \pm 1.27$\\    			
		\bottomrule	
    \end{tabular}
    \label{tab:table_texture_performance}
\end{table}

\subsection{Texture Classification}
\label{sec:exp_texture_rec}

For texture classification, we used the Kylberg dataset~\cite{Kylberg_Dataset} that contains 28 texture classes of different natural and man-made surfaces. 
Each class has 160 unique samples imaged with and without rotation.
Samples from this dataset are shown in Fig.~\ref{fig:kylberg_dataset}.

As in Section~\ref{sec:exp_material_cat}, we resized the images to {$128 \times 128$} pixels
and generated CovDs from 1024 observations obtained on a coarse grid.
The feature vector at each pixel on this grid was taken as
\begin{equation*}
	\Vec{x}_{u,v} =	\bigg[I_{u,v}, \left| \frac{\partial I}  {\partial u}  \right|,  ~\left|\frac{\partial I}  {\partial v}  \right|,
    ~\left| \frac{\partial^2 I}{\partial u^2}\right|,  ~\left|\frac{\partial^2 I}{\partial v^2}\right|~\bigg]^T\;.
\end{equation*}%
We randomly selected 5 images in each class for training and used the remaining ones as test data. 

In Table~\ref{tab:table_texture_performance}, we report recognition accuracies averaged over 10 such random partitions. As before, NN on infinite-dimensional CovDs clearly outperforms NN on $5 \times 5$-dimensional CovDs. Interestingly, it even outperforms the more involved CDL method. With kernel SVM, the accuracies of our  $J_\mathcal{H}$ and 
$S_\mathcal{H}$ divergences are improved to 91\%.

\subsection{Person Re-identification}
\label{sec:exp_person_rec}

For person re-identification, we used two sequences from the ETHZ dataset~\cite{ETHZ_Dataset}.
Sequence~1 contains 83 pedestrians in 4,857 images, and Sequence~2 contains 35 pedestrians in 1,936 images.
We resized all images to {$48 \times 24$} pixels, and, at each pixel $\Vec{u} = (u,v)$, computed the 17-dimensional feature vector
\noindent
\begin{equation*}
\Vec{x}_{\Vec{u}}
\mbox{=}
\big [
  \Vec{u},~
  r_{\Vec{u}},~   g_{\Vec{u}},~   b_{\Vec{u}},~
  \dot r_{\Vec{u}},~  \dot g_{\Vec{u}},~  \dot b_{\Vec{u}},~
  \ddot r_{\Vec{u}},~ \ddot g_{\Vec{u}},~  \ddot b_{\Vec{u}}~
\big]^T\;,
\end{equation*}%
\noindent
where $r_{\Vec{u}}$, $g_{\Vec{u}}$ and $b_{\Vec{u}}$ are the color intensities, and, \eg, for the $r$ channel,
$\dot r_{\Vec{u}} \mbox{=} \big( \left|{\partial r} \middle/ {\partial u}\right|, \left|{\partial r} \middle/ {\partial v}\right| \big)$
and
$\ddot r_{\Vec{u}} \mbox{=} \big( \left|{\partial^2 r} \middle/ {\partial u^2}\right|, \left|{\partial^2 r} \middle/ {\partial v^2}\right| \big)$.
Following~\cite{Bazzani_CVIU_2013}, we randomly selected 10 images from each subject for training and used the rest for testing.

In Table~\ref{tab:table_ethz_performance}, we report the accuracies averaged over 10 random partitions. In addition to the usual baselines, we report the
 state-of-the-art results obtained with the Symmetry-Driven Accumulation of Local Features (SDALF) of~\cite{Bazzani_CVIU_2013}. Once again,
both \textbf{$J_\mathcal{H}$-NN} and \textbf{$S_\mathcal{H}$-NN} 
outperform \textbf{$J$-NN} and \textbf{$S$-NN}, and similarly for SVM.
More importantly, \textbf{$J_\mathcal{H}$-NN} and \textbf{$S_\mathcal{H}$-NN} outperform SDALF, and even more so with kernel SVM.
In supplementary material, we provide the Cumulative Matching Characteristic (CMC) curves that are commonly used for person re-identification.

\begin{table}[!tb]
 	\caption    {    
		   Recognition accuracies for the ETHZ dataset~\cite{ETHZ_Dataset}.
    }
  	\centering
    \begin{tabular}{lcc}
    	\toprule
    	{\bf Method} &{\bf Seq\#1} &{\bf Seq\#2}\\
    	\toprule  	
    	{\bf $J$-NN}									&$80.7\% \pm 1.5$			&$77.2\% \pm 1.2$\\	
    	{\bf $S$-NN}									&$81.3\% \pm 1.5$			&$77.9\% \pm 1.1$\\
    	{\bf SDALF~\cite{Bazzani_CVIU_2013}}   			&$83.4\% \pm N/A$			&$83.4\% \pm N/A$\\
    	{\bf $J$-SVM}									&$83.4\% \pm 1.0$			&$83.1\% \pm 1.2$\\
    	{\bf $S$-SVM}									&$84.4\% \pm 1.0$			&$84.2\% \pm 1.3$\\
    	\midrule
    	{\bf $J_{\mathcal{H}}$-NN}     					&$85.7\% \pm 1.7$			&$84.3\% \pm 1.8$\\
    	{\bf $S_{\mathcal{H}}$-NN}                  	&$85.9\% \pm 1.7$			&$84.5\% \pm 1.8$\\
    	\midrule
    	{\bf $J_{\mathcal{H}}$-SVM}	               		&$89.1\% \pm 1.1$			&$90.9\% \pm 1.1$\\    			
		{\bf $S_{\mathcal{H}}$-SVM}	               		&$\bf 90.2\% \pm 1.0$		&$\bf 91.4\% \pm 0.8$\\    			
		\bottomrule	
    \end{tabular}
    \label{tab:table_ethz_performance}
\end{table}

\subsection{Action Recognition from Motion Capture Data}
\label{sec:exp_action_rec}

Finally, we performed an experiment on human action recognition from motion capture sequences using the HDM05 database~\cite{HDM05_Doc}, which contains 14 different actions. 
Each action is represented by the 3D locations of 31 joints over time. In our experiments, we only used the 4 joints corresponding to arms and legs. This let us compute a 12-dimensional feature vector per frame by concatenating the 3D locations of these 4 joints in that frame. 
The CovDs are then computed over the frames.
We used a leave-one-subject-out setup, where 4 out of the 5 available subjects were used for training and the remaining one for testing. 

In Table~\ref{tab:table_MoCap_performance}, we report the average accuracies over the 5 runs. 
Again, infinite-dimensional CovDs outperform the ones computed from the original observations and yield the best results when used in conjunction with kernel SVM.

\begin{table}[!t]
	\small
 	\caption    {    
		   Recognition accuracies for the HDM05 database~\cite{HDM05_Doc}.
    }
   \vspace{-0.2cm}
  	\centering
    \begin{tabular}{lc}
    	\toprule
    	{\bf Method} &{\bf Recognition Accuracy }\\
    	\toprule  	
    	{\bf $J$-NN}									&$47.3\% \pm 7.0$\\	
    	{\bf $S$-NN}									&$47.8\% \pm 7.4$\\
    	{\bf CDL~\cite{Wang_CVPR_2012_CDL}}    			&$65.3\% \pm 8.9$\\
    	{\bf $J$-SVM}									&$50.8\% \pm 8.4$\\
    	{\bf $S$-SVM}									&$56.8\% \pm 11.5$\\
    	\midrule
    	{\bf $J_{\mathcal{H}}$-NN}     					&$63.3\% \pm 9.4$\\
    	{\bf $S_{\mathcal{H}}$-NN}                  	&$65.9\% \pm 12.8$\\
    	\midrule
    	{\bf $J_{\mathcal{H}}$-SVM}	               		&$70.8\% \pm 8.1$\\    			
		{\bf $S_{\mathcal{H}}$-SVM}	               		&$\bf 73.3\% \pm 11.4$\\    			
		\bottomrule	
    \end{tabular}
    \label{tab:table_MoCap_performance}
\vspace{-0.1cm}
\end{table}

\section{Conclusions and Future Work}
\label{sec:conclusion}

We have introduced an approach to computing infinite-dimensional CovDs, as well as several Bregman divergences to compare them. 
Our experimental evaluation has demonstrated that the resulting infinite-dimensional CovDs lead to state-of-the art recognition accuracies on several challenging datasets.
In the future, we intend to explore how other types of similarity measures, such as the AIRM, can be computed over infinite-dimensional CovDs. Furthermore, we are interested in studying how the Fr\'{e}chet mean of a set of infinite-dimensional CovDs can be evaluated. This would allow us to perform clustering, and would therefore pave the way to extending well-known methods, such as bag of words, to infinite dimensional CovDs.

\appendix 
\section{Appendix}
\label{app:supp_material}

In the following, we provide the detailed derivation of the Bregman divergences in RKHS considered in Section~\textcolor{red}{4} of the main paper. We also
provide the CMC curves for the person re-identification experiment of Section~\textcolor{red}{5.4}, which were left out of the main paper due to space limitation.

\section{Bregman Divergences on RKHS}
Recall that in Section~\textcolor{red}{4} of the main paper, we have exploited the equivalence $\Mat{W}_{\Mat{X}}^T \Phi_{\Mat{X}}^T \Phi_{\Mat{X}} \Mat{W}_{\Mat{X}}  = \Lambda_{\Mat{X}} - \rho \mathbf{I}_\Mat{X}$ (Eq.~\textcolor{red}{12}) to derive Bregman divergences in RKHS. We prove this equivalence below:
\begin{align}
	\Mat{W}_{\Mat{X}}^T \Phi_{\Mat{X}}^T \Phi_{\Mat{X}} \Mat{W}_{\Mat{X}}  &= 
	\Big( \mathbf{I}_{\Mat{X}} - \rho \Lambda_{\Mat{X}}^{-1} \Big)^{1/2} 
	\Mat{V}_{\Mat{X}}^T \Mat{J}_{\Mat{X}}^T 
	\Phi_{\Mat{X}}^T \Phi_{\Mat{X}} \Mat{J}_{\Mat{X}} \Mat{V}_{\Mat{X}} 
	\Big( \mathbf{I}_{\Mat{X}} - \rho \Lambda_{\Mat{X}}^{-1} \Big)^{1/2} \nonumber \\
	&= \Big( \mathbf{I}_{\Mat{X}} - \rho \Lambda_{\Mat{X}}^{-1} \Big)^{1/2}  
	\Mat{V}_{\Mat{X}}^T \Mat{J}_{\Mat{X}}^T \Mat{K}_{\Mat{X},\Mat{X}}\Mat{J}_{\Mat{X}} \Mat{V}_{\Mat{X}}
	\Big( \mathbf{I}_{\Mat{X}} - \rho \Lambda_{\Mat{X}}^{-1} \Big)^{1/2} \nonumber \\
	&= \Big( \mathbf{I}_{\Mat{X}} - \rho \Lambda_{\Mat{X}}^{-1} \Big)^{1/2}  
	\Lambda_{\Mat{X}}\Big( \mathbf{I}_{\Mat{X}} - \rho \Lambda_{\Mat{X}}^{-1} \Big)^{1/2} \nonumber \\
	&= \Lambda_{\Mat{X}} - \rho \mathbf{I}_\Mat{X}.
	\label{eqn:wkwt}
\end{align}
We now provide additional details for the specific Bregman divergences considered in the paper.
The Euclidean metric in RKHS can be derived as
\begin{align}
	\delta_e^2(\widehat{\Mat{C}}_\Mat{X},\widehat{\Mat{C}}_\Mat{Y}) &= 
	\big \| \Phi_{\Mat{X}} \Mat{W}_{\Mat{X}} \Mat{W}_{\Mat{X}}^T \Phi_{\Mat{X}}^T - 
	\Phi_{\Mat{Y}} \Mat{W}_{\Mat{Y}} \Mat{W}_{\Mat{Y}}^T \Phi_{\Mat{Y}}^T \big \|_F^2 \notag \\
	&= \big \| \Phi_{\Mat{X}} \Mat{W}_{\Mat{X}} \Mat{W}_{\Mat{X}}^T \Phi_{\Mat{X}}^T \big{\|}_F^2 + 
   \big{\|}\Phi_{\Mat{Y}} \Mat{W}_{\Mat{Y}} \Mat{W}_{\Mat{Y}}^T \Phi_{\Mat{Y}}^T\big{\|}_F^2 
   -2\tr\big ( 
   \Phi_{\Mat{X}} \Mat{W}_{\Mat{X}} \Mat{W}_{\Mat{X}}^T \Phi_{\Mat{X}}^T 
   \Phi_{\Mat{Y}} \Mat{W}_{\Mat{Y}} \Mat{W}_{\Mat{Y}}^T \Phi_{\Mat{Y}}^T 
   \big ) \notag \\
	&=  \big{\|}\Lambda_\Mat{X} - \rho \mathbf{I}_\Mat{X}\big{\|}_F^2 +   
	\big{\|}\Lambda_\Mat{Y} - \rho \mathbf{I}_\Mat{Y}\big{\|}_F^2  
	-2\tr\big (\Mat{W}_{\Mat{Y}}^T \Mat{K}_{\Mat{Y},\Mat{X}}\Mat{W}_{\Mat{X}} \Mat{W}_{\Mat{X}}^T
	\Mat{K}_{\Mat{X},\Mat{Y}}\Mat{W}_{\Mat{Y}} 
	\big )	
	\notag \\
	&=  \big{\|}\Lambda_\Mat{X} - \rho \mathbf{I}_\Mat{X}\big{\|}_F^2 +   
	\big{\|}\Lambda_\Mat{Y} - \rho \mathbf{I}_\Mat{Y}\big{\|}_F^2  
	-2\big{\|}\Mat{W}_{\Mat{Y}}^T \Mat{K}_{\Mat{Y},\Mat{X}}\Mat{W}_{\Mat{X}}\big{\|}_F^2 .		
	\label{eqn:k_euclidean_metric}
\end{align}

For the Burg and related divergences (\ie, Jeffreys and Stein divergences), we first show that 
$ \det \big( \widehat{\Mat{C}}_{\Mat{X}} \big) = \rho^{\mathcal{|H|}} \det \big( \rho^{-1} \Lambda_{\Mat{X}} \big)$.
To this end, we use the Sylvester determinant theorem, which states that, for two matrices $\Mat{A}$ and $\Mat{B}$ of size $n \times m$ and 
$m \times n$, $\det\Big(\mathbf{I}_n + \Mat{A}\Mat{B} \Big) =   \det\Big(\mathbf{I}_m + \Mat{B}\Mat{A} \Big)$. Therefore,

\begin{align}
	\det \big( \widehat{\Mat{C}}_{\Mat{X}} \big) &= \det \big(\Phi_{\Mat{X}} \Mat{W}_{\Mat{X}} \Mat{W}_{\Mat{X}}^T \Phi_{\Mat{X}}^T
	 + \rho \mathbf{I}_\mathcal{|H|} \big) \notag \\
	&= \rho^\mathcal{|H|}\det \big( \mathbf{I}_{\Mat{X}} + 
	\frac{1}{\rho} \Mat{W}_{\Mat{X}}^T \Phi_{\Mat{X}}^T \Phi_{\Mat{X}} \Mat{W}_{\Mat{X}} \big) \notag \\
	&= \rho^\mathcal{|H|} \det \big(\mathbf{I}_{\Mat{X}} + \frac{1}{\rho} \left( \Lambda_{\Mat{X}} - 
	\rho \mathbf{I}_{\Mat{X}} \right) \big) \notag \\
	&= \rho^{\mathcal{|H|}} \det \big( \rho^{-1} \Lambda_{\Mat{X}} \big).	
	\label{eqn:det_cov_rkhs}
\end{align}

We then make use of the Woodbury identity, which states that, for two matrices $\Mat{A}$ and $\Mat{B}$ of size $n \times m$ and $m \times n$, 
\begin{equation*}
\Big(\Mat{A}\Mat{B} + \rho \mathbf{I}_n \Big)^{-1} = \frac{1}{\rho}\mathbf{I}_n - \frac{1}{\rho^2}\Mat{A}
\Big(\mathbf{I}_m + \frac{1}{\rho}\Mat{B}\Mat{A}\Big)^{-1}\Mat{B}. 
\end{equation*}
This lets us write
\begin{align}
	\widehat{\Mat{C}}^{-1}_{\Mat{Y}} &= \Big( \Phi_{\Mat{Y}} \Mat{W}_{\Mat{Y}} \Mat{W}_{\Mat{Y}}^T \Phi_{\Mat{Y}}^T + 
	\rho \mathbf{I}_\mathcal{|H|} \Big)^{-1} \notag \\
	&= \frac{1}{\rho} \mathbf{I}_\mathcal{|H|} - \frac{1}{\rho^2} \Phi_{\Mat{Y}} \Mat{W}_{\Mat{Y}} 
	\Big(\mathbf{I}_\Mat{Y} + \frac{1}{\rho}\Mat{W}_{\Mat{Y}}^T \Phi_{\Mat{Y}}^T \Phi_{\Mat{Y}} \Mat{W}_{\Mat{Y}}\Big)^{-1}	
	\Mat{W}_{\Mat{Y}}^T \Phi_{\Mat{Y}}^T \notag \\
	&= \frac{1}{\rho} \mathbf{I}_\mathcal{|H|} - \frac{1}{\rho^2} \Phi_{\Mat{X}} \Mat{W}_{\Mat{Y}} 
	\Big(\mathbf{I}_\Mat{Y} + \frac{1}{\rho}\big(\Lambda_\Mat{Y} - \rho\mathbf{I}_\Mat{Y}\big) \Big)^{-1}	
	\Mat{W}_{\Mat{Y}}^T \Phi_{\Mat{Y}}^T \notag \\
	&= \frac{1}{\rho} \mathbf{I}_\mathcal{|H|} - \frac{1}{\rho} \Phi_{\Mat{Y}} \Mat{W}_{\Mat{Y}} 
	\Lambda_\Mat{Y}^{-1} \Mat{W}_{\Mat{Y}}^T \Phi_{\Mat{Y}}^T.
	\label{eqn:p_inv_CX}
\end{align}

Therefore, we have
\begin{align}
	\tr\Big(\widehat{\Mat{C}}_{\Mat{X}}\widehat{\Mat{C}}^{-1}_{\Mat{Y}}\Big) &=
		\tr\Big(
		\big( \Phi_{\Mat{X}} \Mat{W}_{\Mat{X}} \Mat{W}_{\Mat{X}}^T \Phi_{\Mat{X}}^T	+\rho \mathbf{I}_\mathcal{|H|} \big)
		\rho^{-1}\big(\mathbf{I}_\mathcal{|H|} - 
		\Phi_{\Mat{Y}} \Mat{W}_{\Mat{Y}} \Lambda_\Mat{Y}^{-1} \Mat{W}_{\Mat{Y}}^T \Phi_{\Mat{Y}}^T\big)
		 \Big)
		\notag \\
	&=	\tr (\mathbf{I}_\mathcal{|H|}) + \rho^{-1}\tr \big( \Phi_{\Mat{X}} \Mat{W}_{\Mat{X}} \Mat{W}_{\Mat{X}}^T \Phi_{\Mat{X}}^T \big)
	    -\tr \big( \Phi_{\Mat{Y}} \Mat{W}_{\Mat{Y}} \Lambda_\Mat{Y}^{-1} \Mat{W}_{\Mat{Y}}^T \Phi_{\Mat{Y}}^T \big)
	    \notag \\
	&	\hspace{3ex}-\rho^{-1}\tr \Big( \Phi_{\Mat{X}} \Mat{W}_{\Mat{X}} \Mat{W}_{\Mat{X}}^T \Phi_{\Mat{X}}^T
		\Phi_{\Mat{Y}} \Mat{W}_{\Mat{Y}} \Lambda_\Mat{Y}^{-1} \Mat{W}_{\Mat{Y}}^T \Phi_{\Mat{Y}}^T 
	      \Big)
	     \notag \\
	&=	\mathcal{|H|} + \rho^{-1}\tr \big( \Lambda_{\Mat{X}} - \rho \mathbf{I}_\Mat{X} \big)
	    -\tr \big( \mathbf{I}_\Mat{Y}  - \rho \Lambda_{\Mat{Y}}^{-1} \big)
	    -\rho^{-1}\tr \Big( \Mat{W}_{\Mat{X}}^T \Mat{K}_{\Mat{X},\Mat{Y}} \Mat{W}_{\Mat{Y}} \Lambda_\Mat{Y}^{-1} \Mat{W}_{\Mat{Y}}^T 
	    \Mat{K}_{\Mat{Y},\Mat{X}} \Mat{W}_{\Mat{X}} \Big).
	    \notag \\	     
	\label{eqn:tr_Xinv_Y}
\end{align}

Recall from Definition~\textcolor{red}{2.3} that the Burg divergence between to SPD matrices can be written as
\begin{align}
		B(\Mat{C}_1,\Mat{C}_2) &=
		\tr (\Mat{C}_1\Mat{C}^{-1}_2) -\ldet\big(\Mat{C}_1\Mat{C}_2^{-1}\big) -n\;.
\end{align}

Using Eq.~\ref{eqn:det_cov_rkhs} and Eq.~\ref{eqn:tr_Xinv_Y}, we can thus derive the Burg divergence in RKHS as 
\begin{align}
	B_{\mathcal{H}}\Big(\widehat{\Mat{C}}_{\Mat{X}},\widehat{\Mat{C}}_{\Mat{Y}}\Big) &=
		\frac{1}{\rho}\tr \big( \Lambda_{\Mat{X}} - \rho \mathbf{I}_\Mat{X} \big)
	    -\tr \big( \mathbf{I}_\Mat{Y}  - \rho \Lambda_{\Mat{Y}}^{-1} \big) \notag \\
	    &-\frac{1}{\rho}\tr \Big( \Mat{W}_{\Mat{X}}^T \Mat{K}_{\Mat{X},\Mat{Y}} \Mat{W}_{\Mat{Y}} \Lambda_\Mat{Y}^{-1} \Mat{W}_{\Mat{Y}}^T 
	    \Mat{K}_{\Mat{Y},\Mat{X}} \Mat{W}_{\Mat{X}} \Big)	    	     
	    + \ldet\Big(\rho^{-1} \Lambda_{\Mat{Y}}\Big) - \ldet\Big(\rho^{-1} \Lambda_{\Mat{X}}\Big). 
	\label{eqn:burg_rkhs}
\end{align}

%
Having the Burg divergence at our disposal, it is straightforward to obtain the Jeffreys divergence, which is given by
$\frac{1}{2}B_{\mathcal{H}}\big(\widehat{\Mat{C}}_{\Mat{X}},\widehat{\Mat{C}}_{\Mat{Y}}\big) + 
\frac{1}{2}B_{\mathcal{H}}\big(\widehat{\Mat{C}}_{\Mat{Y}},\widehat{\Mat{C}}_{\Mat{X}}\big)$. This divergence can thus be written as
\begin{align}
	2J_{\mathcal{H}}\Big(\widehat{\Mat{C}}_{\Mat{X}},\widehat{\Mat{C}}_{\Mat{Y}}\Big) &=
		\frac{1}{\rho}\tr \big( \Lambda_{\Mat{Y}} - \rho \mathbf{I}_\Mat{Y} \big)
	    -\tr \big( \mathbf{I}_\Mat{X}  - \rho \Lambda_{\Mat{X}}^{-1} \big) \notag \\
	    &-\frac{1}{\rho}\tr \Big( \Mat{W}_{\Mat{Y}}^T \Mat{K}_{\Mat{Y},\Mat{X}} \Mat{W}_{\Mat{X}} \Lambda_\Mat{X}^{-1} \Mat{W}_{\Mat{X}}^T 
	    \Mat{K}_{\Mat{X},\Mat{Y}} \Mat{W}_{\Mat{Y}} \Big)	    	     
	    + \ldet \Big(\rho^{-1} \Lambda_{\Mat{Y}}\Big) - \ldet \Big(\rho^{-1} \Lambda_{\Mat{X}}\Big)
	    \notag \\
		&+\frac{1}{\rho}\tr \big( \Lambda_{\Mat{X}} - \rho \mathbf{I}_\Mat{X} \big)
	    -\tr \big( \mathbf{I}_\Mat{Y}  - \rho \Lambda_{\Mat{Y}}^{-1} \big) \notag \\
	    &-\frac{1}{\rho}\tr \Big( \Mat{W}_{\Mat{X}}^T \Mat{K}_{\Mat{X},\Mat{Y}} \Mat{W}_{\Mat{Y}} \Lambda_\Mat{Y}^{-1} \Mat{W}_{\Mat{Y}}^T 
	    \Mat{K}_{\Mat{Y},\Mat{X}} \Mat{W}_{\Mat{X}} \Big)	    	     
	    + \ldet\Big(\rho^{-1} \Lambda_{\Mat{X}}\Big) - \ldet \Big(\rho^{-1} \Lambda_{\Mat{Y}}\Big)	
	    \notag\\    
	    &=\frac{1}{\rho}\tr \big( \Lambda_{\Mat{X}} - \rho \mathbf{I}_\Mat{X} \big) + 
	    \frac{1}{\rho}\tr \big( \Lambda_{\Mat{Y}} - \rho \mathbf{I}_\Mat{Y} \big) 
	    -\tr \big( \mathbf{I}_\Mat{X}  - \rho \Lambda_{\Mat{X}}^{-1} \big)  -
	    \tr \big( \mathbf{I}_\Mat{Y}  - \rho \Lambda_{\Mat{Y}}^{-1} \big) \notag\\
	    &-\frac{1}{\rho}\tr \Big( \Mat{W}_{\Mat{X}}^T \Mat{K}_{\Mat{X},\Mat{Y}} \Mat{W}_{\Mat{Y}} \Lambda_\Mat{Y}^{-1} \Mat{W}_{\Mat{Y}}^T 
	    \Mat{K}_{\Mat{Y},\Mat{X}} \Mat{W}_{\Mat{X}} \Big)  
	    -\frac{1}{\rho}\tr \Big( \Mat{W}_{\Mat{Y}}^T \Mat{K}_{\Mat{Y},\Mat{X}} \Mat{W}_{\Mat{X}} \Lambda_\Mat{X}^{-1} \Mat{W}_{\Mat{X}}^T 
	    \Mat{K}_{\Mat{X},\Mat{Y}} \Mat{W}_{\Mat{Y}} \Big). 
	\label{eqn:jefferys_rkhs}
\end{align}

The details of the derivation of the Stein divergence are already given in the main paper. We therefore omit this divergence here.

\end{document}